\documentclass[9pt,sigconf,screen=true,anonymous=false,bookmarks=false,nonacm]{acmart}
\setcopyright{none}
\usepackage{enumitem}
\setlist{leftmargin=5.08mm}

\usepackage{lipsum}
\usepackage{graphicx}
\usepackage{amsmath}
\usepackage{footnote}
\usepackage{mathtools}
\usepackage{mathrsfs}     
\usepackage{comment}
\usepackage{soul}                                          
\usepackage[subrefformat=parens,labelformat=parens]{subfig}
\usepackage{bm,bbm}
\usepackage{multirow}
\usepackage{threeparttable,booktabs}
\usepackage{blkarray}
\usepackage{tikz}
\usetikzlibrary{positioning,calc,fit,decorations.pathmorphing,shapes.geometric,shapes.gates.logic.US,calc}
\usetikzlibrary{arrows,arrows.meta,decorations.markings,shapes,shapes.arrows}
\usetikzlibrary{patterns,snakes}
\usetikzlibrary{decorations,decorations.pathreplacing}
\usetikzlibrary{backgrounds}
\usepackage{tikz-3dplot}
\usepackage{balance}
\usepackage{courier}                                       
\usepackage{cleveref}                                      
\usepackage[mathcal]{eucal}
\usepackage[]{algpseudocode}                               
\algrenewcommand\textproc{\texttt}
\makeatletter\let\float@addtolists\relax\makeatother
\usepackage{algorithm}
\usepackage{filecontents}                                  
\usepackage{pgfplots}
\usepackage{pgfplotstable}
\usepackage{pgf-pie}
\usepgfplotslibrary{groupplots}
\pgfplotsset{compat=newest}
\usepackage[figuresright]{rotating}
\usepackage{xcolor,colortbl}                               
\usepackage{adjustbox}                                     
\usepackage{makecell}
\usepackage{siunitx}
\usepackage{color}
\usepackage{diagbox}

\newcommand{\minisection}[1]{\vspace{.06in}\noindent{\textbf{#1}}.}

\theoremstyle{plain}

\theoremstyle{definition}

\algrenewcommand\textproc{\texttt}

\usepackage[skip=1pt]{caption}            
\definecolor{CUHKorange}{RGB}{244,106,18} 
\definecolor{CUHKblue}{RGB}{0,111,190}    
\definecolor{CUHKgreen}{RGB}{0,127,128}   
\definecolor{CUHKred}{RGB}{228,46,36}     
\definecolor{CUHKyellow}{RGB}{198,148,34} 
\definecolor{CUHKdark}{RGB}{114,44,114}   
\definecolor{CUHKmiddle}{RGB}{144,44,144} 

\usepackage{tcolorbox}
\tcbuselibrary{skins,breakable}
    {\endtcolorbox}
    {\endtcolorbox}

\graphicspath{{./figs/}}

\usepackage{listings}
\usepackage{array}

\begin{document}

\title{Verilog-Evolve: Feedback-Driven and Skill-Evolving Verilog Generation}

\author{
    Zehua Pei$^1$, 
    Hui-Ling Zhen$^2$,
    Yu Zhang$^1$,
    Sinno Jialin Pan$^1$,
    Mingxuan Yuan$^2$,
    Bei Yu$^1$\\
$^1$The Chinese University of Hong Kong \quad
$^2$Huawei Technologies Co., Ltd
}

\begin{abstract}

Large language models (LLMs) have improved Verilog generation from natural-language specifications, but most pipelines still treat generation as isolated sampling followed by functional checking.
This is insufficient for practical RTL design, where useful Verilog must be correct, synthesizable, timing-conscious, and friendly to downstream hardware objectives.
We present \textbf{Verilog-Evolve}, a feedback-driven framework for versioned Verilog refinement and cross-session skill evolution.
For each task, Verilog-Evolve generates diverse minor candidates, evaluates them with executable feedback from functional simulation, Yosys synthesis, ABC timing proxy, and optional GEMM metrics, then promotes the best candidate into a major version under configurable scoring.
To improve across tasks, the system maintains modular skill guidance, retrieves skills according to task and feedback context, and evolves candidate skills from logged histories through create/improve/skip decisions and verifier reports.
Experiments on VerilogEval and mixed-precision GEMM tasks show that Verilog-Evolve improves final functional success and promotion stability while producing more downstream-friendly RTL under open-source synthesis, timing-proxy, and netlist-level GEMM objectives.
Validation-gated skill evolution further improves GEMM downstream quality and achieves the best downstream score and GEMM held-out pass rate among the evaluated skill modes.

\end{abstract}

\maketitle
\pagestyle{plain}

\section{Introduction}

Large language models (LLMs) have shown strong potential for hardware description language (HDL) generation, including Verilog generation from natural-language specifications~\cite{fan2023large, nam2024using, pei2024betterv}. 
However, Verilog is not ordinary software: it is an executable hardware blueprint whose usefulness depends not only on textual plausibility, but also on functional correctness, synthesizability, timing behavior, and downstream hardware cost~\cite{liu2023chipnemo, he2023chateda}. 
As LLM-generated RTL moves closer to electronic design automation (EDA) workflows, a code snippet that merely resembles Verilog, or even passes a limited functional test, is insufficient for practical hardware design~\cite{pu2024customized, zhang2024sola}.

Existing LLM-based Verilog generation methods largely optimize for one-shot correctness or pass@k on functional benchmarks~\cite{zhang2024mg, pei2024betterv, thakur2024verigen, liu2023rtlcoder, lu2023rtllm, gao2024autovcoder, ho2025verilogcoder, yao2024hdldebugger}. 
This setting hides three limitations.
First, model-only generation and self-reflection provide weak evidence: an LLM may assert that a module satisfies a specification while the implementation still fails compilation or simulation.
Second, functional correctness alone does not ensure downstream quality; two functionally equivalent Verilog implementations can differ substantially in synthesized cell count, wire complexity, timing proxy, and suitability for accelerator kernels.
Third, most generation pipelines treat each attempt as an isolated sample, without a versioned mechanism to record tool outcomes, promote better candidates, or accumulate reusable repair knowledge across tasks.

Recent work offers two complementary perspectives beyond isolated functional sampling.
RTL optimization methods use LLM-guided rewriting, symbolic reasoning, long-context design context, and tool feedback to improve existing RTL designs under synthesis or timing objectives~\cite{yao2024rtlrewriter, wang2026symrtlo, ye2026longrtl, fang2026drrtl}.
SkillClaw, on the other hand, shows that agent skills can be evolved from cross-session traces through aggregation, create/improve/skip decisions, validation queues, and worker-side replay~\cite{skillclaw2026}.
Together, these directions motivate a generation process that closes the loop with tools, versions candidate artifacts, and converts successful traces into reusable skills.
We therefore formulate Verilog generation as a feedback-driven refinement problem rather than a single-pass translation problem.
The key idea is to close the loop between LLM generation and executable feedback: each candidate is evaluated by concrete tools, scored by configurable objectives, and either repaired or superseded by a better candidate.
Unlike pure RTL optimization, Verilog-Evolve starts from a natural-language specification and required module declaration rather than a fixed golden RTL.
Unlike generic skill-evolution systems, its memory is grounded in HDL-specific evidence, including simulation outcomes, synthesis statistics, timing proxies, downstream GEMM metrics, and optional SEC/PPA reports.

\begin{figure}[htbp]
    \centering
    \includegraphics[width=0.7\linewidth]{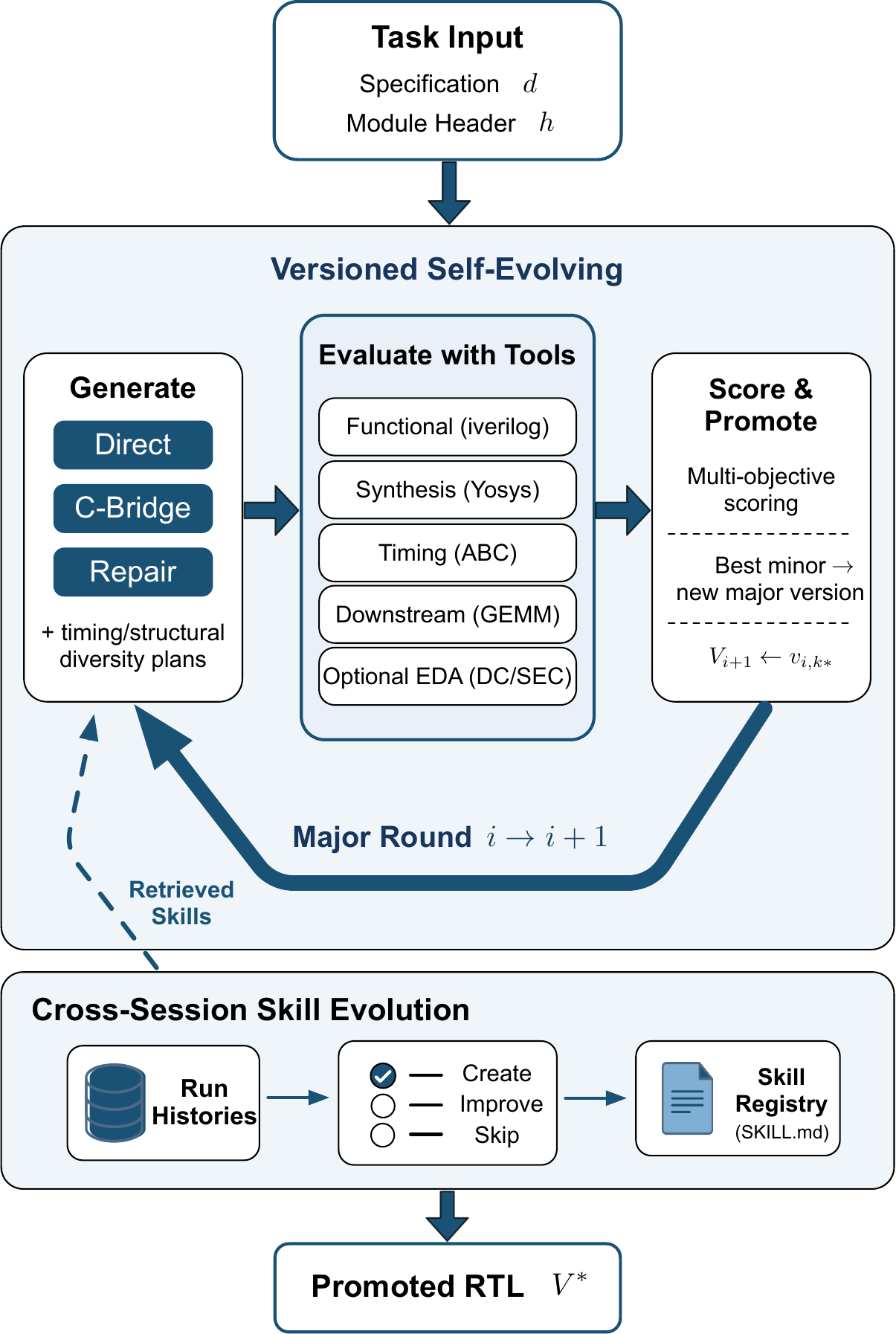} 
    \caption{
    Overview of Verilog-Evolve. Given a natural-language specification and module header, the framework performs task-level major/minor versioned search: multiple skill-conditioned minor RTL candidates are generated, evaluated by executable tool feedback, scored under configurable objectives, and promoted into the next major version. Cross-session histories are further distilled into verifier- and validation-gated skill updates, which are retrieved to guide future generation and repair.
    }
    \label{fig:overview}
\end{figure}

To realize this formulation, we introduce \textbf{Verilog-Evolve}, a feedback-driven framework that combines versioned search with skill guidance, illustrated in \Cref{fig:overview}.
For each design task, Verilog-Evolve launches multiple \emph{minor} candidates using generation strategies such as direct Verilog generation, C-bridged generation, and feedback-conditioned repair.
When timing or synthesis feedback is available, each minor can further use path- or module-level hints to diversify candidate generation.
Each candidate is evaluated by a pluggable evaluator pool: a functional evaluator based on \texttt{iverilog}/\texttt{vvp}, a Yosys synthesis evaluator, an ABC-based timing proxy, a downstream evaluator for mixed-precision GEMM-style objectives, and an optional DC/SEC-style EDA adapter for industrial PPA evaluation.
The evaluator outputs are converted into a multi-objective score, enabling the system to select the best minor candidate and promote it into the next \emph{major} version when it improves the current baseline and passes the configured gate.
This major/minor protocol lets Verilog-Evolve refine candidates across versions.

Beyond task-level search, Verilog-Evolve maintains a persistent skill memory.
Reusable guidance is stored as modular skill files and retrieved according to task descriptions, timing paths, diversity plans, and skill registry history.
After a batch of runs, structured histories are ingested as sessions, aggregated by referenced skills, and converted into \emph{create}, \emph{improve}, or \emph{skip} decisions.
Candidate skills are checked by evidence-based verifiers and can either be published immediately or queued for external replay validation workers before entering the persistent skill registry.
This design allows the system to improve both within a task, through feedback-driven candidate promotion, and across tasks, through validation-gated skill evolution.

The main contributions of this paper are:
\begin{itemize}
    \item We formulate specification-to-RTL generation as a \emph{feedback-driven refinement} problem, using a major/minor loop to evaluate, repair, and promote candidates.
    
    \item We develop a pluggable evaluator and configurable scoring framework that supports functional simulation, Yosys synthesis metrics, ABC timing proxy, downstream-aware GEMM objectives, and optional industrial DC/SEC feedback.
        
    \item We introduce timing-plan-guided candidate diversification and cross-session, validation-gated skill evolution for Verilog generation.
    
    \item Experimental results on VerilogEval and mixed-precision GEMM tasks show that Verilog-Evolve preserves functional correctness while improving synthesis, timing-proxy, and task-specific hardware metrics.
\end{itemize}

\section{Related Work}

\subsection{LLM-based Verilog Generation and Optimization}
Large language models have been increasingly explored for Verilog and RTL generation~\cite{thakur2023benchmarking, chang2023chipgpt, pei2024betterv}.
To address data scarcity, prior work constructs or augments RTL corpora from open-source designs~\cite{dehaerne2023deep, liu2023rtlcoder, thakur2024verigen, cui2024origen}.
BetterV transfers general code knowledge to Verilog through discriminative guidance~\cite{pei2024betterv}, while MG-Verilog and CodeV build high-quality instruction data by summarizing and transforming RTL designs~\cite{zhang2024mg, zhao2024codev}.
Recent reasoning-oriented work such as CodeV-R1 further improves generation through reinforcement learning~\cite{zhu2025codev}.
VerilogEval and RTLLM provide functional-correctness benchmarks~\cite{liu2023verilogeval, lu2023rtllm}, later extended by VerilogEval-V2~\cite{pinckney2024revisitingverilogevalnewerllms}.
Agentic methods further decompose generation and debugging: AutoVCoder uses systematic prompting~\cite{gao2024autovcoder}; VerilogCoder introduces planning and waveform-tracing tools~\cite{ho2025verilogcoder}; and MAGE employs a multi-agent engine~\cite{zhao2024mage}. 
RTLFixer~\cite{tsai2023rtlfixer} and HDLDebugger~\cite{yao2024hdldebugger} use compiler or simulator outputs to repair generated RTL, but primarily target syntactic validity or functional correctness.
LLM-aided RTL optimization typically starts from an existing RTL design and uses rewriting, symbolic guidance, long-context analysis, and timing-driven feedback to improve it~\cite{yao2024rtlrewriter, wang2026symrtlo, ye2026longrtl, fang2026drrtl}.

\subsection{Self-Evolving Agents and Memory}
A growing line of work studies agents that improve future behavior from execution traces.
General multi-agent frameworks such as AutoGen, AutoAgents, and LAMBDA coordinate multiple LLM agents for complex tasks~\cite{wu2023autogen, chen2023autoagents, sun2024lambda}, while other paradigms use self-evaluation, reflection, or prompt evolution to adapt behavior over time~\cite{agentcollab, rethinker, pei2025scope}.
SkillClaw provides a concrete skill-memory design: it aggregates cross-session trajectories, groups evidence by referenced skills, decides whether to create, improve, or skip a skill update, and optionally gates publication through replay validation workers~\cite{skillclaw2026,lin2025towards}.
Verilog-Evolve adapts this mechanism to RTL generation, grounding skill updates in HDL tool outcomes rather than conversational traces. 

In contrast to prior work, Verilog-Evolve combines several ideas into a unified specification-to-RTL framework.
Unlike one-shot generators, it organizes search as versioned major/minor refinement with self-evolving skills.
Unlike tool-feedback systems that only repair syntax or functional errors, it scores and selects candidates under configurable objectives using a multi-fidelity evaluator stack that includes functional simulation, Yosys synthesis~\cite{Yosys}, ABC timing proxy~\cite{ABC}, and downstream GEMM metrics.
Unlike generic skill-evolution systems, its skill memory is grounded in HDL-specific evidence, and unlike existing RTL optimizers that start from golden designs, it begins from natural-language specifications.

\section{Method}

Verilog-Evolve treats Verilog generation as iterative candidate refinement driven by executable feedback.
Given a task description $\mathbf{d}$ and a required module declaration $\mathbf{h}$, the system repeatedly generates candidate implementations, evaluates them with executable tools, scores them under configurable objectives, promotes the best candidates, and evolves reusable skills from histories.
This section covers skill retrieval, diverse minor attempts, the evaluator pipeline, scoring, promotion, and cross-session skill evolution.

\begin{algorithm}[t]
\caption{Verilog-Evolve}
\label{alg:verilog-evolve}
\begin{algorithmic}[1]
\Require task $\mathbf{d}$, module header $\mathbf{h}$, skills $\mathcal{S}$, evaluators $\mathcal{E}$
\Ensure best RTL $\mathbf{V}^*$ and updated skills
\State Initialize baseline $\mathbf{V}_0$ and feedback context $\mathcal{C}_0$;
\For{$i=0$ to $R-1$}
    \For{$k=1$ to $K$}
        \State Sample strategy $g_{i,k}$ and diversity plan $p_{i,k}$;
        \State Retrieve skills $\mathcal{S}_{i,k}\leftarrow \textit{Retrieve}(\mathcal{S},\mathbf{d},\mathcal{C}_i,p_{i,k})$;
        \State Generate/repair $\mathbf{v}_{i,k}\leftarrow \textit{LLM}(g_{i,k},\mathbf{d},\mathbf{h},\mathcal{S}_{i,k},\mathcal{C}_i)$;
        \State Evaluate $r_{i,k}\leftarrow \{E_j(\mathbf{v}_{i,k})\}_{j=1}^{m}$ and log trajectory;
        \If{early-stop mode is enabled and $\textit{Pass}(r_{i,k})$}
            \State break minor loop;
        \EndIf
    \EndFor
    \State $\mathcal{A}_i\leftarrow\{k:\textit{Pass}(r_{i,k})\}$;
    \If{$\mathcal{A}_i\neq\emptyset$}
        \State $k^*\leftarrow \arg\min_{k\in\mathcal{A}_i} \big(\textit{Score}(r_{i,k}),\textit{Tie}(r_{i,k})\big)$;
    \EndIf
    \If{$\mathcal{A}_i\neq\emptyset$ and $\textit{Improve}(i,k^*)$ and $\textit{Gate}(r_{i,k^*},\mathbf{v}_{i,k^*})$}
        \State $\mathbf{V}_{i+1}\leftarrow \mathbf{v}_{i,k^*}$; update $\mathcal{C}_{i+1}$;
    \Else
        \State $\mathbf{V}_{i+1}\leftarrow \mathbf{V}_i$; carry forward $\mathcal{C}_{i+1}\leftarrow \mathcal{C}_i$;
    \EndIf
\EndFor
\State Evolve skills from logged sessions via create/improve/skip and verifier gates;
\State Publish immediately or queue validated skills for replay workers;
\end{algorithmic}
\end{algorithm}

\subsection{Skill Retrieval and Prompt Conditioning}

Verilog-Evolve uses a persistent skill memory to guide both generation and repair.
Each skill is stored as a lightweight skill file containing trigger conditions and reusable domain guidance.
The current skill library covers functional Verilog generation, simulator-feedback repair, synthesis-aware rewriting, timing-aware rewriting, RTL optimization patterns, and downstream co-design for quantized GEMM-style kernels.
For each minor attempt, the system retrieves relevant skills according to three sources of context: the natural-language task, current timing or synthesis feedback, and the sampled diversity plan.
The retriever also consults the skill registry, so skills with validated version histories receive a higher retrieval score.
Each candidate record stores whether its prompt used static or retrieved skills, enabling later skill attribution.

Formally, let $\mathcal{S}_{i,k}$ denote the retrieved skill subset for minor attempt $k$ in major round $i$, and let $\mathbf{p}_{i,k}$ denote its diversity plan.
The generation context is $\mathbf{x}_{i,k}=(\mathbf{d},\mathbf{h},\mathcal{S}_{i,k},\mathbf{p}_{i,k})$.
The generator produces a candidate Verilog implementation by conditioning on this context:
\begin{equation}
    \mathbf{v}_{i,k} = \textit{LLM}_{gen}(\mathbf{d}, \mathbf{h}, \mathcal{S}_{i,k}, \mathbf{p}_{i,k}).
\end{equation}
When a previous candidate fails, the repair prompt additionally receives structured tool feedback $\mathbf{f}$:
\begin{equation}
    \mathbf{v}'_{i,k} = \textit{LLM}_{repair}(\mathbf{d}, \mathbf{h}, \mathbf{v}, \mathbf{f}, \mathcal{S}_{i,k}, \mathbf{p}_{i,k}).
\end{equation}
Skill guidance does not replace tool evaluation; it provides reusable priors that help the model produce more synthesizable, timing-friendly, and downstream-aware RTL.

\subsection{Diverse Minor Attempts}

Verilog-Evolve organizes generation into major rounds and minor attempts.
In each major round $i$, the system launches $K$ minor candidates $\{\mathbf{v}_{i,1},\ldots,\mathbf{v}_{i,K}\}$ using two complementary diversity sources.

\minisection{Generation Strategy}
The first source is the generation strategy.
Direct generation asks the LLM to generate a complete Verilog module from the task description, module declaration, and retrieved skills.
The C-bridge strategy first asks for a C-style functional reference, then uses it to guide Verilog generation.
Feedback-conditioned repair starts from the current major candidate and applies structured feedback from failed evaluators.
For example, a functional mismatch may trigger reset, signedness, or bit-width repair hints; a synthesis failure may trigger simplification of unsupported constructs.

\minisection{Timing and Structural Diversity Plan}
The second source is an optional timing- and structure-aware diversity plan.
When timing paths or synthesis feedback are available, the system may use path- or module-level hints to diversify minor attempts.
These hints can prioritize timing-critical regions, structurally complex modules, or random exploration targets, while the rewrite focus rotates among combinational, sequential, and mixed transformations.
Thus, a minor attempt is characterized by
\begin{equation}
    \mathbf{v}_{i,k} \sim 
    \textit{Strategy}_{i,k}
    \times
    \textit{PathSelect}_{i,k}
    \times
    \textit{Focus}_{i,k}.
\end{equation}
\subsection{Executable Evaluator Pipeline}

Each minor candidate is evaluated by a pluggable evaluator pool.
Let $\mathcal{E}=\{E_1,\ldots,E_m\}$ be the enabled evaluator set.
For a candidate $\mathbf{v}$, each evaluator returns
\begin{equation}
    r_j = E_j(\mathbf{v}) = (\textit{pass}_j, \mathbf{m}_j, \mathbf{f}_j),
\end{equation}
where $\textit{pass}_j$ is the evaluator status, $\mathbf{m}_j$ is a metric dictionary, and $\mathbf{f}_j$ is textual feedback used for repair.
For each candidate $\mathbf{v}_{i,k}$, the full evaluator record is the collection of all enabled evaluator outputs:
\begin{equation}
    r_{i,k} = \mathcal{E}(\mathbf{v}_{i,k})
    = \{r_{i,k,j}\}_{j=1}^{m},\quad
    r_{i,k,j}=E_j(\mathbf{v}_{i,k}).
\end{equation}
Thus the $r_j$ notation denotes a single evaluator output, while $r_{i,k}$ denotes the complete candidate-level record consumed by scoring, repair, and promotion.
The scalar score is computed only from $\textit{pass}_{i,k,j}$ and metrics $\mathbf{m}_{i,k,j}$; feedback $\mathbf{f}_{i,k,j}$ is used to condition later repair prompts.

\minisection{Functional Evaluator}
The functional evaluator uses \texttt{iverilog} and \texttt{vvp} through the VerilogEval harness to compile and simulate a candidate.
It classifies outcomes into \textit{passed}, \textit{mismatch}, \textit{compile\_error}, \textit{syntax\_error}, \textit{timeout}, or \textit{unknown\_failure}.
For mismatch failures, it records the number of mismatched samples and total samples, and converts common failure patterns into repair hints.

\minisection{Yosys Synthesis Evaluator}
The synthesis evaluator invokes Yosys to parse, elaborate, optimize, and report structural statistics~\cite{Yosys}.
It extracts metrics such as cell count, wire count, wire bits, and cell-type distribution.
These metrics allow the search loop to prefer simpler and more synthesis-friendly RTL among functionally correct candidates.

\minisection{ABC Timing Proxy}
The timing evaluator uses Yosys' ABC mapping flow as a portable proxy for timing quality~\cite{ABC}.
It reports logic-cell count, DFF count, and an ABC delay proxy.
While this proxy is not a replacement for full static timing analysis, it provides a lightweight downstream signal for discouraging long or overly complex combinational structures.

\minisection{Downstream Evaluator}
For downstream-aware experiments, Verilog-Evolve supports task-specific evaluators.
In our mixed-precision GEMM setting, the downstream evaluator checks whether generated RTL is friendly to mixed-precision GEMM kernels.
It reports features such as multiplier count, adder count, DFF count, mux count, preferred bit-width hits, pipeline-depth proxies, area-delay-product proxies, and an aggregate downstream score.

\minisection{Optional Industrial EDA Evaluator}
When an industrial flow is available, Verilog-Evolve can invoke or parse a DC/SEC-style evaluator.
This evaluator returns SEC status, WNS, TNS, area, power, and critical paths.
SEC is treated as a hard correctness gate for promotion in this mode.
The implementation also supports a parse-only mode, allowing the framework to consume pre-existing industrial EDA reports without requiring commercial tools on every machine.

\minisection{Held-Out Promotion Evaluator}
For selected downstream tasks, visible tests are used for repair feedback while generated held-out tests are used only for major-version promotion.
In the mixed-precision GEMM benchmark, the held-out evaluator generates randomized cases for supported tasks and rejects candidates that overfit the visible testbench.

\subsection{Configurable Multi-Objective Scoring}

The evaluator outputs are merged by a configurable scoring function.
For a candidate record $r$, scoring returns both a scalar objective and a Boolean eligibility bit:
\begin{equation}
    \textit{EvalScore}(r)=(\textit{Score}(r),\textit{Pass}(r)).
\end{equation}
Let $\mathcal{R}$ be the evaluator names marked as required by the score configuration.
The Boolean pass flag is
\begin{equation}
    \textit{Pass}(r)=
    \bigwedge_{E_j\in\mathcal{R}}\textit{pass}_j
    \wedge
    \bigwedge_{g\in\mathcal{G}_{hard}} g(r),
\end{equation}
where $\mathcal{G}_{hard}$ contains explicit hard gates such as SEC success.
Required evaluators and hard gates determine promotion eligibility, while optional evaluator failures contribute only soft penalties unless the configuration promotes them into $\mathcal{R}$ or $\mathcal{G}_{hard}$.
The functional evaluator is required in all reported runs and acts as the primary correctness gate: candidates that fail compilation, simulation, or timeout receive large penalties and are ineligible for promotion under the default configuration.
For open-source evaluation, the score is computed as
\begin{equation}
    \textit{Score}_{\textit{open}}(r) =
    P_{\textit{func}}(r)
    + P_{\textit{opt}}(r)
    + \lambda_a A(r)
    + \lambda_w W(r)
    + \lambda_t T(r)
    + \lambda_d D(r),
\end{equation}
where $P_{\textit{func}}$ is the functional penalty, $P_{\textit{opt}}$ is the fixed penalty for failed optional evaluators, $A$ and $W$ denote normalized synthesis area and wiring proxies, $T$ denotes the ABC timing proxy, and $D$ denotes the downstream task score.
The weights are selected by a score configuration, allowing correctness-only, PPA-aware, timing-aware, or downstream-aware runs.
A mismatch receives a penalty proportional to its mismatch ratio, while compile errors, syntax errors, timeouts, and unknown failures receive fixed penalties.
Optional evaluators can either contribute soft penalties or be marked as required by the score configuration; $P_{\textit{opt}}$ accumulates a fixed penalty for each enabled optional evaluator that fails.
In implementation, the metric terms are obtained by merging metric dictionaries from the enabled evaluator records.
The synthesis terms $A(r)$ and $W(r)$ are computed from Yosys metrics such as \texttt{cell\_count}, \texttt{wire\_count}, and \texttt{wire\_bits}; $T(r)$ is computed from ABC metrics such as \texttt{abc\_delay\_proxy}; and $D(r)$ is computed from downstream-evaluator metrics such as \texttt{downstream\_score} and GEMM structural proxies.
The score configuration specifies which metric keys are weighted and normalized, so additional evaluator metrics can be included without changing the scoring interface.
Yosys, ABC, and downstream evaluators can be configured either as soft objectives or as required evaluators.

Verilog-Evolve also supports an industrial-style scoring mode aligned with RTL timing optimization:
\begin{equation}
    \textit{Score}_{\textit{EDA}}(r) =
    0.5 \cdot \Delta_{\textit{WNS}}
    + 0.35 \cdot \Delta_{\textit{TNS}}
    + 0.15 \cdot \Delta_{\textit{Area}}
    + P_{\textit{area}} + P_{\textit{SEC}},
\end{equation}
where deltas are normalized against the baseline major version.
If SEC fails, $P_{\textit{SEC}}$ makes the candidate ineligible for promotion.
This mode separates open-source proxy evaluation from industrial PPA evaluation: ABC delay is used only as a lightweight proxy, while WNS/TNS/area and SEC are used when the EDA adapter is enabled.

\subsection{Versioned Candidate Promotion}

Verilog-Evolve promotes candidates through a major/minor versioning protocol.
During major round $i$, the system evaluates $K$ minor candidates $\mathbf{v}_{i,1},\ldots,\mathbf{v}_{i,K}$.
Let $\mathcal{A}_i$ be the eligible set:
\begin{equation}
    \mathcal{A}_i =
    \{k \mid \textit{Pass}(r_{i,k})\}.
\end{equation}
SEC is enforced through the hard-gate component of $\textit{Pass}(r)$ when the industrial EDA mode is active.
Candidates outside $\mathcal{A}_i$ may still be useful for feedback and skill evidence, but they cannot become the next major version.
Among eligible candidates, the system selects the lowest score using a deterministic tie-break rule:
\begin{equation}
    k^* = \arg\min_{k\in\mathcal{A}_i}
    \left(\textit{Score}(r_{i,k}),\textit{Tie}(r_{i,k})\right).
\end{equation}
The default $\textit{Tie}$ is the minor index, so exact score ties are resolved by the earlier generated candidate.
If $\mathcal{A}_i$ is empty, no candidate is promoted in that round.
Otherwise, the selected candidate is considered an improvement iff its scalar score is strictly lower than the current major baseline score:
\begin{equation}
    \textit{Improve}(i,k^*) \equiv
    \textit{Score}(r_{i,k^*}) <
    \textit{Score}(r_{\textit{major},i}).
\end{equation}
For the first major version, the improvement predicate is true for any eligible candidate.
When a promotion-only gate such as held-out functional testing is enabled, it is applied to the selected minor before the major-version update.
If no promotion-only gate is enabled, $\textit{Gate}(r_{i,k^*},\mathbf{v}_{i,k^*})$ defaults to true.
The update rule is therefore
\begin{equation}
    \mathbf{V}_{i+1} \leftarrow
    \begin{cases}
    \mathbf{v}_{i,k^*}, & \mathcal{A}_i\neq\emptyset
    \wedge \textit{Improve}(i,k^*) \wedge \textit{Gate}(r_{i,k^*},\mathbf{v}_{i,k^*})\\
    \mathbf{V}_i, & \text{otherwise}.
    \end{cases}
\end{equation}
For PPA-aware and downstream-aware modes, the system evaluates the full minor set and compares candidates directly under the configured objective.
In non-PPA runs, candidate generation can stop as soon as a functionally passing minor is found.
Each minor attempt is recorded in a line-based history file, while each major round records the selected minor, score, promotion decision, and promoted artifact.
The evaluator record of each promoted major version is retained as $r_{\textit{major},i}$ for later improvement checks.

\subsection{Cross-Session Skill Evolution}

After a batch of runs, Verilog-Evolve ingests histories as sessions and drains them into a local evolver store.
Each session is summarized with metadata such as referenced skills, SEC/functional pass count, average score, strategy, path-selection mode, and optimization focus.
Sessions are then aggregated by referenced skill; a single session may contribute to multiple skill groups.
For each group, the evolver first applies an evidence gate.
Let $n_{\textit{pass}}$ be the number of records that pass the active correctness gate, $n_{\textit{promote}}$ be the number of records that provide promotion-gate or promoted-major evidence, and $\overline{\Delta s}$ be the mean logged score-improvement signal.
The verifier accepts a candidate skill only when $n_{\textit{pass}}\ge \tau_{\textit{pass}} \wedge (n_{\textit{promote}}\ge \tau_{\textit{promote}} \vee \overline{\Delta s}<0)$, 
    with defaults $\tau_{\textit{pass}}=2$ and $\tau_{\textit{promote}}=1$.
If the candidate has high equivalence risk, at least one promotion is required even when the average score improves.
An accepted group updates an existing matching skill through \emph{improve\_skill}; if no matching skill exists, it becomes \emph{create\_skill}. Groups that fail the evidence gate are assigned \emph{skip}.

\minisection{Immediate and Validated Publication}
Verilog-Evolve supports two publication modes.
In immediate mode, verifier-approved skills are written back to the corresponding skill file and registered with a stable skill identifier and version hash.
In validated mode, the candidate skill is placed into a validation queue with replay cases, current skill text, candidate skill text, and publication thresholds.
External validation workers replay up to a bounded set of cases with both baseline and candidate skill guidance, evaluate the generated RTL, and append one validator result per worker.
Each result contains an approval bit and a quality score in $[0,1]$, where a passing replay receives quality 1 and failing replays are mapped from their scalar evaluation score.
In the default implementation, the current worker replays at most three cases per job and uses the functional evaluator for this replay check; the queue itself may store up to eight candidate replay cases for independent workers.
The default queue requirements are $N_{\textit{results}}\ge 2$, $N_{\textit{approve}}\ge 2$ and $\overline{q}_{\textit{cand}}\ge 0.75$. 
A worker approves only if the candidate skill's mean replay quality is no worse than the baseline skill and also exceeds the queue's minimum score threshold.
The publisher aggregates worker outputs by counting approvals and averaging reported candidate quality scores.
If the required number of results has not arrived, the job remains pending.
If enough results have arrived but either the approval count or mean score threshold is not met, the job is rejected.
Only jobs satisfying all three thresholds are published and registered.

In summary, Verilog-Evolve improves within a task through feedback-driven version promotion and across tasks through validation-gated skill evolution.

\section{Experiments}
This section evaluates whether Verilog-Evolve improves Verilog generation without sacrificing correctness, and whether the additional tool- and skill-evolution components provide stable gains under a fixed maximum search budget.
We organize the study around three questions:
\begin{itemize}
    \item \textbf{RQ1: Functional stability.} Does versioned repair and promotion preserve or improve functional correctness compared with direct generation, C-bridge generation, and repair-only baselines?
    \item \textbf{RQ2: Tool-guided selection.} Do open-source synthesis and timing proxies help select more stable final candidates without relying on commercial EDA tools?
    \item \textbf{RQ3: Skill evolution.} Does cross-session skill retrieval and validation-gated skill evolution further improve final correctness and promotion stability?
\end{itemize}

\begin{table*}[htbp]
    \centering
    \caption{\small Functional correctness and promotion stability on VerilogEval.
    All evaluated methods use the same maximum search budget: 3 major rounds and 5 minor candidates per round; Verilog-Evolve variants use the strategy pool \texttt{direct,c\_bridge,repair}.
    ``Final Success'' reports whether the final selected candidate passes the public functional tests.
    ``Promotion Pass'' reports the fraction of selected candidates that satisfy the configured promotion gate.
    Lower functional score is better.}
    \label{tab:main_functional}
    \renewcommand{\arraystretch}{1.12}
    \resizebox{\linewidth}{!}{
    \begin{tabular}{|l|c|cc|cc|cc|c|c|}
        \hline
        \multirow{2}{*}{Method}
        & \multirow{2}{*}{Tool Feedback}
        & \multicolumn{2}{c|}{VerilogEval-Machine}
        & \multicolumn{2}{c|}{VerilogEval-Human}
        & \multicolumn{2}{c|}{Stability}
        & \multirow{2}{*}{Avg. Promoted}
        & \multirow{2}{*}{Skill Val. A/R} \\
        \cline{3-8}
        &
        & Final Success $\uparrow$ & Best Func. Score $\downarrow$
        & Final Success $\uparrow$ & Best Func. Score $\downarrow$
        & Promotion Pass $\uparrow$ & Compile Pass $\uparrow$
        & & \\
        \hline
        Direct LLM
        & none
        & 76.8 & 0.46
        & 61.5 & 0.71
        & 58.9 & 84.2
        & 0.0
        & -- \\

        C-Bridge
        & none
        & 79.4 & 0.41
        & 64.7 & 0.64
        & 61.8 & 86.5
        & 0.0
        & -- \\

        Repair-only
        & \texttt{iverilog}
        & 81.2 & 0.35
        & 67.1 & 0.57
        & 64.9 & 88.1
        & 1.1
        & -- \\

        Verilog-Evolve w/o Skill Evolution
        & \texttt{iverilog}+\texttt{yosys}+\texttt{abc}
        & 83.6 & 0.29
        & 70.4 & 0.49
        & 68.7 & 90.3
        & 1.8
        & -- \\

        Verilog-Evolve Full
        & \texttt{iverilog}+\texttt{yosys}+\texttt{abc}+skills
        & \textbf{85.3} & \textbf{0.23}
        & \textbf{72.8} & \textbf{0.41}
        & \textbf{71.6} & \textbf{91.9}
        & \textbf{2.2}
        & 18/6 \\
        \hline
    \end{tabular}
    }
\end{table*}

\begin{table*}[htbp]
    \centering
    \caption{\small Open-source multi-objective evaluation on mixed-precision GEMM tasks.
    F, Y, A, and D denote functional simulation, Yosys synthesis proxy, ABC timing proxy, and downstream GEMM evaluator, respectively.
    Mul Cells, DFF Cells, and ADP Proxy are parsed from Yosys JSON netlists by the downstream evaluator.
    Lower cell count, multiplier usage, ABC delay proxy, ADP proxy, and downstream score are better.}
    \label{tab:gemm_downstream}
    \renewcommand{\arraystretch}{1.08}
    \resizebox{\linewidth}{!}{
    \begin{tabular}{|l|c|cc|cccccc|}
        \hline
        Variant & Eval. Stack
        & Func. Pass $\uparrow$ & GEMM Held-out Pass $\uparrow$
        & Cell Count $\downarrow$ & Mul Cells $\downarrow$
        & DFF Cells & ABC Delay $\downarrow$
        & ADP Proxy $\downarrow$ & Downstream Score $\downarrow$ \\
        \hline
        Correctness-only
        & F
        & 91.7 & 83.3
        & 182.4 & 4.7
        & 18.3 & 6.81
        & 512.6 & 4.92 \\
        PPA-aware
        & F+Y
        & 91.7 & 86.1
        & 159.8 & 3.8
        & 16.9 & 6.43
        & 428.4 & 4.31 \\
        Timing-aware
        & F+Y+A
        & 88.9 & 86.1
        & 164.2 & 3.9
        & 19.6 & 5.72
        & 401.7 & 4.08 \\
        Downstream-aware
        & F+Y+A+D
        & \textbf{91.7} & \textbf{88.9}
        & \textbf{151.6} & \textbf{2.4}
        & 21.8 & \textbf{5.48}
        & \textbf{332.5} & \textbf{3.21} \\
        \hline
    \end{tabular}
    }
\end{table*}

\subsection{Experimental Setting}

\minisection{Benchmark and Model}
We use VerilogEval-Machine and VerilogEval-Human~\cite{liu2023verilogeval} as the functional-correctness benchmark.
Unless otherwise stated, all variants use BetterV-CodeQwen-7B as the base model.
This choice focuses the evaluation on the algorithmic contribution of Verilog-Evolve rather than model scaling. 

\minisection{Tools and Objectives}
The main experiments use only open-source tools.
The functional evaluator compiles and simulates candidates with iverilog.
The synthesis evaluator uses Yosys to extract structural proxies such as cell count and wire bits.
The timing evaluator uses Yosys' ABC integration as a lightweight timing proxy.
For downstream GEMM experiments, the downstream evaluator measures task-specific structural features such as multiplier usage, DFF count, mux count, pipeline-depth proxy, and area-delay-product proxy.
The optional DC/SEC adapter is not used in the main open-source results unless explicitly stated.

\minisection{Search Budget and Variants}
All Verilog-Evolve variants use the same maximum search budget: 3 major rounds, 5 minor candidates per round, and the strategy pool direct,c\_bridge,repair.
We compare five variants.
\textbf{Direct LLM} samples direct Verilog without tool feedback.
\textbf{C-Bridge} first asks for a C-style reference and then generates Verilog.
\textbf{Repair-only} uses iverilog feedback for local repair but does not use multi-tool versioned selection.
\textbf{Verilog-Evolve w/o Skill Evolution} uses versioned major/minor search and open-source tool feedback, but disables cross-session skill evolution.
\textbf{Verilog-Evolve Full} enables skill retrieval, skill evolution, verifier gates, and validation-aware skill publication.

\minisection{Metrics}
For VerilogEval, we report final success on public functional tests, best functional score, promotion pass rate, compile pass rate, average number of promoted major versions, and skill validation accepted/rejected counts.
Final success measures whether the final selected candidate passes the public VerilogEval tests.
Promotion pass measures whether selected candidates satisfy the configured promotion criteria under the enabled evaluator stack.
The functional score is lower when a candidate has fewer or less severe failures.

\subsection{Functional Correctness and Promotion Stability}
Table~\ref{tab:main_functional} reports the main functional-correctness result on VerilogEval.

The results show consistent improvement from direct generation to the full Verilog-Evolve pipeline.
C-bridge generation improves over direct generation, indicating that an intermediate software-like representation is helpful for clarifying datapath behavior.
Repair-only further improves both final success and compile pass rate, showing that simulator feedback provides useful local correction signals.
Adding versioned multi-tool selection improves final success from 81.2 to 83.6 on VerilogEval-Machine and from 67.1 to 70.4 on VerilogEval-Human.

The full system achieves the best final success, the lowest functional score, and the highest promotion pass rate.
Compared with the versioned system without skill evolution, it improves the final success from 83.6 to 85.3 (VerilogEval-Machine) and from 70.4 to 72.8 (VerilogEval-Human).
The promotion pass rate increases from 68.7 to 71.6, and the average number of promoted major versions rises to 2.2.
Finally, the 18/6 accepted/rejected skill-validation ratio shows that skill updates are filtered by verifier and validation gates.
These results support RQ1: feedback-driven versioned repair preserves and improves functional correctness compared with direct generation, C-bridge generation, and repair-only baselines.

\subsection{Downstream-Aware GEMM Evaluation}

Table~\ref{tab:gemm_downstream} evaluates whether the open-source evaluator stack improves downstream hardware quality on the mixed-precision GEMM benchmark.
This experiment uses the three built-in GEMM-style tasks: int4\_int8\_mac\_pe, mixed\_precision\_dot4, and requantize\_int32\_to\_int8.
We compare four evaluator configurations under the same maximum search budget.
\textbf{Correctness-only} uses only functional simulation.
\textbf{PPA-aware} adds Yosys structural metrics.
\textbf{Timing-aware} further adds ABC timing proxy.
\textbf{Downstream-aware} adds the GEMM-specific downstream evaluator, which parses Yosys JSON netlists to expose multiplier cells, register cells, and an area-delay-product proxy.
The reported metrics are averaged over the three GEMM tasks.

The results show that functional correctness alone is insufficient to select GEMM-friendly RTL.
Correctness-only generation reaches a high functional pass rate, but produces larger structures and worse downstream score.
Adding Yosys and ABC reduces the cell count and delay proxy while improving downstream score.
The downstream-aware variant achieves the best overall hardware quality: it matches the best functional pass rate, improves GEMM held-out promotion pass rate to 88.9, and obtains the lowest cell count, multiplier count, ABC delay proxy, ADP proxy, and downstream score.
This supports RQ2: open-source tool feedback can guide Verilog-Evolve toward downstream-friendly RTL without relying on commercial DC/SEC flows.

\subsection{Effect of Skill Evolution}

Table~\ref{tab:skill_evolution} studies the contribution of skill retrieval, cross-session evolution, and validation-gated publication.
We compare four modes.
\textbf{No skills} disables skill guidance entirely.
\textbf{Static skills} uses the manually written skill files but does not update them.
\textbf{Evolved, immediate} publishes verifier-approved skills directly after a run.
\textbf{Evolved, validated} queues candidate skills for external replay workers before publishing them.

\begin{table}[t]
    \centering
    \caption{\small Effect of skill evolution and validation.
    Human Success is the final functional success on VerilogEval-Human.
    GEMM Score and GEMM Held-out Pass are both measured only on the mixed-precision GEMM downstream tasks.
    GEMM Score is the average downstream score, and GEMM Held-out Pass is the randomized GEMM promotion-test pass rate.
    Skill A/R reports accepted/rejected skill updates.}
    \label{tab:skill_evolution}
    \renewcommand{\arraystretch}{1.08}
    \resizebox{\linewidth}{!}{
    \begin{tabular}{|l|c|c|c|c|}
        \hline
        Skill Mode
        & Human Success $\uparrow$
        & GEMM Score $\downarrow$
        & GEMM Held-out Pass $\uparrow$
        & Skill A/R \\
        \hline
        No skills
        & 68.9 & 4.37 & 82.1 & -- \\
        Static skills
        & 70.4 & 3.98 & 85.0 & -- \\
        Evolved, immediate
        & 72.1 & 3.55 & 87.3 & 24/0 \\
        Evolved, validated
        & \textbf{72.8} & \textbf{3.21} & \textbf{88.9} & 18/6 \\
        \hline
    \end{tabular}
    }
\end{table}

Static skills improve over no-skill prompting, indicating that reusable RTL guidance helps both general VerilogEval generation and GEMM-oriented design.
Evolving skills further improves Human success from 70.4 to 72.1, reduces GEMM downstream score from 3.98 to 3.55, and raises the GEMM held-out pass rate from 85.0 to 87.3.
The validated mode publishes fewer updates, accepting 18 and rejecting 6, but achieves the best Human success, the lowest GEMM downstream score, and the highest GEMM held-out pass rate.
Overall, replay-based validation filters skill updates before they enter the shared skill registry, supporting RQ3.

\section{Conclusion}

We presented Verilog-Evolve, a feedback-driven framework for versioned Verilog refinement with cross-session skill evolution.
Experiments on VerilogEval and mixed-precision GEMM tasks show that feedback-driven evaluation and validation-gated skill updates improve both functional correctness and downstream hardware quality.
Future work will extend the evaluator stack with richer physical-design signals and integrate industrial EDA flows with formal equivalence checking.

\clearpage
\bibliographystyle{IEEEtran-sim}
\bibliography{ref/main2, ref/Top, ref/main}

@string{cav      = "International Conference on Computer-Aided Verification (CAV)"}

@string{date     = "IEEE/ACM Proceedings Design, Automation and Test in Eurpoe (DATE)"}

@string{iccad    = "IEEE/ACM International Conference on Computer-Aided Design (ICCAD)"}

@string{mlcad    = "ACM/IEEE Workshop on Machine Learning CAD (MLCAD)"}

@string{aaai     = "AAAI Conference on Artificial Intelligence (AAAI)"}

@string{open     = "Optimization and Engineering"}

@string{arxiv    = "arXiv preprint"}

@article{wang2026symrtlo,
  title={Symrtlo: Enhancing rtl code optimization with llms and neuron-inspired symbolic reasoning},
  author={Wang, Yiting and Ye, Wanghao and Guo, Ping and He, Yexiao and Wang, Ziyao and Tian, Bowei and He, Shwai and Sun, Guoheng and Shen, Zheyu and Chen, Sihan and others},
  journal={Advances in Neural Information Processing Systems},
  volume={38},
  pages={50093--50118},
  year={2026}
}

@article{ye2026longrtl,
  title={LongRTL: Graph-Similarity-Guided LLM-driven Long Context RTL Optimization},
  author={Ye, Yuyang and Shen, Che-Kuan and Hu, Xiangfei and Liu, Yuchen and Yin, Shuo and Yao, Xufeng and Yu, Bei and Ho, Tsung-Yi},
  year={2026}
}

@inproceedings{yao2024rtlrewriter,
  title={Rtlrewriter: Methodologies for large models aided rtl code optimization},
  author={Yao, Xufeng and Wang, Yiwen and Li, Xing and Lian, Yingzhao and Chen, Ran and Chen, Lei and Yuan, Mingxuan and Xu, Hong and Yu, Bei},
  booktitle={Proceedings of the 43rd IEEE/ACM International Conference on Computer-Aided Design},
  pages={1--7},
  year={2024}
}

@inproceedings{ABC,
  title={{ABC}: An academic industrial-strength verification tool},
  author={Brayton, Robert and Mishchenko, Alan},
  booktitle=cav,
  year={2010}
}

@misc{pinckney2024revisitingverilogevalnewerllms,
      title={Revisiting VerilogEval: Newer LLMs, In-Context Learning, and Specification-to-RTL Tasks}, 
      author={Nathaniel Pinckney and Christopher Batten and Mingjie Liu and Haoxing Ren and Brucek Khailany},
      year={2024},
      eprint={2408.11053},
      archivePrefix={arXiv},
      primaryClass={cs.SE},
      url={https://arxiv.org/abs/2408.11053}, 
}

@inproceedings{ho2025verilogcoder,
  title={Verilogcoder: Autonomous verilog coding agents with graph-based planning and abstract syntax tree (ast)-based waveform tracing tool},
  author={Ho, Chia-Tung and Ren, Haoxing and Khailany, Brucek},
  booktitle={Proceedings of the AAAI Conference on Artificial Intelligence},
  volume={39},
  number={1},
  pages={300--307},
  year={2025}
}

@article{zhao2024mage,
  title={Mage: A multi-agent engine for automated rtl code generation},
  author={Zhao, Yujie and Zhang, Hejia and Huang, Hanxian and Yu, Zhongming and Zhao, Jishen},
  journal={arXiv preprint arXiv:2412.07822},
  year={2024}
}

@article{zhu2025codev,
  title={CodeV-R1: Reasoning-Enhanced Verilog Generation},
  author={Zhu, Yaoyu and Huang, Di and Lyu, Hanqi and Zhang, Xiaoyun and Li, Chongxiao and Shi, Wenxuan and Wu, Yutong and Mu, Jianan and Wang, Jinghua and Zhao, Yang and others},
  journal={arXiv preprint arXiv:2505.24183},
  year={2025}
}

@article{sun2024lambda,
  title={LAMBDA: A Large Model Based Data Agent},
  author={Sun, Maojun and Han, Ruijian and Jiang, Binyan and Qi, Houduo and Sun, Defeng and Yuan, Yancheng and Huang, Jian},
  journal={arXiv preprint arXiv:2407.17535},
  year={2024}
}

@article{wu2023autogen,
  title={Autogen: Enabling next-gen llm applications via multi-agent conversation framework},
  author={Wu, Qingyun and Bansal, Gagan and Zhang, Jieyu and Wu, Yiran and Zhang, Shaokun and Zhu, Erkang and Li, Beibin and Jiang, Li and Zhang, Xiaoyun and Wang, Chi},
  journal={arXiv preprint arXiv:2308.08155},
  year={2023}
}

@article{chen2023autoagents,
  title={Autoagents: A framework for automatic agent generation},
  author={Chen, Guangyao and Dong, Siwei and Shu, Yu and Zhang, Ge and Sesay, Jaward and Karlsson, B{\"o}rje F and Fu, Jie and Shi, Yemin},
  journal={arXiv preprint arXiv:2309.17288},
  year={2023}
}

@article{chang2023chipgpt,
  title={ChipGPT: How far are we from natural language hardware design},
  author={Chang, Kaiyan and Wang, Ying and Ren, Haimeng and Wang, Mengdi and Liang, Shengwen and Han, Yinhe and Li, Huawei and Li, Xiaowei},
  journal={arXiv preprint arXiv:2305.14019},
  year={2023}
}

@inproceedings{nam2024using,
  title={Using an llm to help with code understanding},
  author={Nam, Daye and Macvean, Andrew and Hellendoorn, Vincent and Vasilescu, Bogdan and Myers, Brad},
  booktitle={Proceedings of the IEEE/ACM 46th International Conference on Software Engineering},
  pages={1--13},
  year={2024}
}

@inproceedings{fan2023large,
  title={Large language models for software engineering: Survey and open problems},
  author={Fan, Angela and Gokkaya, Beliz and Harman, Mark and Lyubarskiy, Mitya and Sengupta, Shubho and Yoo, Shin and Zhang, Jie M},
  booktitle={2023 IEEE/ACM International Conference on Software Engineering: Future of Software Engineering (ICSE-FoSE)},
  pages={31--53},
  year={2023},
  organization={IEEE}
}

@article{pu2024customized,
  title={Customized Retrieval Augmented Generation and Benchmarking for EDA Tool Documentation QA},
  author={Pu, Yuan and He, Zhuolun and Qiu, Tairu and Wu, Haoyuan and Yu, Bei},
  journal={arXiv preprint arXiv:2407.15353},
  year={2024}
}

@article{zhang2024sola,
  title={DiLA: Enhancing LLM tool learning with differential logic layer},
  author={Zhang, Yu and Zhen, Hui-Ling and Pei, Zehua and Lian, Yingzhao and Yin, Lihao and Yuan, Mingxuan and Yu, Bei},
  booktitle={Proceedings of the 32nd ACM SIGKDD Conference on Knowledge Discovery and Data Mining V. 1},
  pages={1952--1963},
  year={2026}
}

@article{gao2024autovcoder,
  title={AutoVCoder: A Systematic Framework for Automated Verilog Code Generation using LLMs},
  author={Gao, Mingzhe and Zhao, Jieru and Lin, Zhe and Ding, Wenchao and Hou, Xiaofeng and Feng, Yu and Li, Chao and Guo, Minyi},
  journal={arXiv preprint arXiv:2407.18333},
  year={2024}
}

@article{yao2024hdldebugger,
  title={Hdldebugger: Streamlining hdl debugging with large language models},
  author={Yao, Xufeng and Li, Haoyang and Chan, Tsz Ho and Xiao, Wenyi and Yuan, Mingxuan and Huang, Yu and Chen, Lei and Yu, Bei},
  journal={arXiv preprint arXiv:2403.11671},
  year={2024}
}

@article{thakur2024verigen,
  title={Verigen: A large language model for verilog code generation},
  author={Thakur, Shailja and Ahmad, Baleegh and Pearce, Hammond and Tan, Benjamin and Dolan-Gavitt, Brendan and Karri, Ramesh and Garg, Siddharth},
  journal={ACM Transactions on Design Automation of Electronic Systems},
  volume={29},
  number={3},
  pages={1--31},
  year={2024},
  publisher={ACM New York, NY}
}

@article{cui2024origen,
  title={OriGen: Enhancing RTL Code Generation with Code-to-Code Augmentation and Self-Reflection},
  author={Cui, Fan and Yin, Chenyang and Zhou, Kexing and Xiao, Youwei and Sun, Guangyu and Xu, Qiang and Guo, Qipeng and Song, Demin and Lin, Dahua and Zhang, Xingcheng and others},
  journal={arXiv preprint arXiv:2407.16237},
  year={2024}
}

@article{zhao2024codev,
  title={CodeV: Empowering LLMs for Verilog Generation through Multi-Level Summarization},
  author={Zhao, Yang and Huang, Di and Li, Chongxiao and Jin, Pengwei and Nan, Ziyuan and Ma, Tianyun and Qi, Lei and Pan, Yansong and Zhang, Zhenxing and Zhang, Rui and others},
  journal={arXiv preprint arXiv:2407.10424},
  year={2024}
}

@article{zhang2024mg,
  title={MG-Verilog: Multi-grained Dataset Towards Enhanced LLM-assisted Verilog Generation},
  author={Zhang, Yongan and Yu, Zhongzhi and Fu, Yonggan and Wan, Cheng and others},
  journal={arXiv preprint arXiv:2407.01910},
  year={2024}
}

@article{pei2024betterv,
  title={Betterv: Controlled verilog generation with discriminative guidance},
  author={Pei, Zehua and Zhen, Hui-Ling and Yuan, Mingxuan and Huang, Yu and Yu, Bei},
  journal={arXiv preprint arXiv:2402.03375},
  year={2024}
}

@article{agentcollab,
  title={AgentCollab: A Self-Evaluation-Driven Collaboration Paradigm for Efficient LLM Agents},
  author={Gao, Wenbo and Liu, Renxi and Wang, Xian and Guo, Fang and Yang, Shuai and Chen, Xi and Zhen, Hui-Ling and Chen, Hanting and Lin, Weizhe and Li, Xiaosong and others},
  journal={arXiv preprint arXiv:2603.26034},
  year={2026}
}

@article{rethinker,
  title={ReThinker: Scientific Reasoning by Rethinking with Guided Reflection and Confidence Control},
  author={Tang, Zhentao and Cui, Yuqi and Kai, Shixiong and Zhao, Wenqian and Ye, Ke and Li, Xing and Tian, Anxin and Pei, Zehua and Zhen, Hui-Ling and Hu, Shoubo and others},
  journal={arXiv preprint arXiv:2602.04496},
  year={2026}
}

@article{lin2025towards,
  title={Towards Efficient Agents: A Co-Design of Inference Architecture and System},
  author={Lin, Weizhe and Zhen, Hui-Ling and Yang, Shuai and Wang, Xian and Liu, Renxi and Chen, Hanting and Zhang, Wangze and Zhou, Chuansai and Li, Yiming and Chen, Chen and others},
  journal={arXiv preprint arXiv:2512.18337},
  year={2025}
}

@article{pei2025scope,
  title={Scope: Prompt evolution for enhancing agent effectiveness},
  author={Pei, Zehua and Zhen, Hui-Ling and Kai, Shixiong and Pan, Sinno Jialin and Wang, Yunhe and Yuan, Mingxuan and Yu, Bei},
  journal={arXiv preprint arXiv:2512.15374},
  year={2025}
}

@MISC{Yosys,
	author = {Claire Wolf},
	title = {Yosys Open SYnthesis Suite},
	howpublished = "\url{https://yosyshq.net/yosys/}"
}

@inproceedings{he2023chateda,
  title={{ChatEDA: A large language model powered autonomous agent for EDA}},
  author={He, Zhuolun and Wu, Haoyuan and Zhang, Xinyun and Yao, Xufeng and Zheng, Su and Zheng, Haisheng and Yu, Bei},
  booktitle=mlcad,
  pages={1--6},
  year={2023},
  organization={IEEE}
}

@article{liu2023chipnemo,
  title={{ChipNeMo: Domain-Adapted LLMs for Chip Design}},
  author={Liu, Mingjie and Ene, Teodor-Dumitru and Kirby, Robert and Cheng, Chris and Pinckney, Nathaniel and Liang, Rongjian and Alben, Jonah and Anand, Himyanshu and Banerjee, Sanmitra and Bayraktaroglu, Ismet and others},
  journal={arXiv preprint arXiv:2311.00176},
  year={2023}
}

@article{tsai2023rtlfixer,
  title={{RTLFixer: Automatically Fixing RTL Syntax Errors with Large Language Models}},
  author={Tsai, YunDa and Liu, Mingjie and Ren, Haoxing},
  journal={arXiv preprint arXiv:2311.16543},
  year={2023}
}

@article{liu2023rtlcoder,
  title={{RTLCoder: Outperforming GPT-3.5 in Design RTL Generation with Our Open-Source Dataset and Lightweight Solution}},
  author={Liu, Shang and Fang, Wenji and Lu, Yao and Zhang, Qijun and Zhang, Hongce and Xie, Zhiyao},
  journal={arXiv preprint arXiv:2312.08617},
  year={2023}
}

@article{lu2023rtllm,
  title={{RTLLM: An open-source benchmark for design rtl generation with large language model}},
  author={Lu, Yao and Liu, Shang and Zhang, Qijun and Xie, Zhiyao},
  journal={arXiv preprint arXiv:2308.05345},
  year={2023}
}

@inproceedings{liu2023verilogeval,
  title={{VerilogEval: Evaluating Large Language Models for Verilog Code Generation}},
  author={Liu, Mingjie and Pinckney, Nathaniel and Khailany, Brucek and Ren, Haoxing},
  booktitle=iccad,
  pages={1--8},
  year={2023},
  organization={IEEE}
}

@article{dehaerne2023deep,
  title={{A Deep Learning Framework for Verilog Autocompletion Towards Design and Verification Automation}},
  author={Dehaerne, Enrique and Dey, Bappaditya and Halder, Sandip and De Gendt, Stefan},
  journal={arXiv preprint arXiv:2304.13840},
  year={2023}
}

@inproceedings{thakur2023benchmarking,
  title={{Benchmarking Large Language Models for Automated Verilog RTL Code Generation}},
  author={Thakur, Shailja and Ahmad, Baleegh and Fan, Zhenxing and Pearce, Hammond and Tan, Benjamin and Karri, Ramesh and Dolan-Gavitt, Brendan and Garg, Siddharth},
  booktitle=date,
  pages={1--6},
  year={2023},
  organization={IEEE}
}

@article{fang2026drrtl,
  title={Dr. RTL: Autonomous Agentic RTL Optimization through Tool-Grounded Self-Improvement},
  author={Fang, Wenji and Lu, Yao and Liu, Shang and Wang, Jing and Guo, Ziyan and He, Junxian and Tu, Fengbin and Xie, Zhiyao},
  journal={arXiv preprint arXiv:2604.14989},
  year={2026}
}

@article{skillclaw2026,
  title={SkillClaw: Let Skills Evolve Collectively with Agentic Evolver},
  author={Ji, Yuxiang and others},
  journal={arXiv preprint arXiv:2604.08377},
  year={2026}
}

\end{document}